# Federated Learning for Predictive Maintenance and Quality Inspection in Industrial Applications

Viktorija Pruckovskaja
*Technische Universität Wien*
Vienna, Austria
viktorija.pruckovskaja
@gmail.com

Axel Weissenfeld
*Data Science & Artificial Intelligence*
*AIT Austrian Institute of Technology*
Vienna, Austria
axel.weissenfeld@ait.ac.at

Clemens Heistracher
*Data Science & Artificial Intelligence*
*AIT Austrian Institute of Technology*
Vienna, Austria
clemens.heistracher@ait.ac.at

Anita Graser
*Data Science & Artificial Intelligence*
*AIT Austrian Institute of Technology*
Vienna, Austria
anita.graser@ait.ac.at

Julia Kafka
*Innovation Software*
*Stiwa Automation*
Attnang-Puchheim, Austria
julia.kafka@stiwa.com

Peter Leputsch
*Innovation Software*
*Stiwa Automation*
Attnang-Puchheim, Austria
Peter.Leputsch@stiwa.com

Daniel Schall
*Distributed AI Systems Research*
*Siemens Technology*
Vienna, Austria
Daniel.Schall@Siemens.com

Jana Kemnitz
*Distributed AI Systems Research*
*Siemens Technology*
Vienna, Austria
Jana.Kemnitz@Siemens.com

*Abstract*—Data-driven machine learning is playing a crucial role in the advancements of Industry 4.0, specifically in enhancing predictive maintenance and quality inspection. Federated learning (FL) enables multiple participants to develop a machine learning model without compromising the privacy and confidentiality of their data. In this paper, we evaluate the performance of different FL aggregation methods and compare them to central and local training approaches. Our study is based on four datasets with varying data distributions. The results indicate that the performance of FL is highly dependent on the data and its distribution among clients. In some scenarios, FL can be an effective alternative to traditional central or local training methods. Additionally, we introduce a new federated learning dataset from a real-world quality inspection setting.

## I. INTRODUCTION

More and more production companies employ data-driven machine learning (ML) models during the production process for predictive maintenance [1] or quality inspection [2]. Data-driven ML approaches depend heavily on the availability of training data. Tracking the data generated by machines, production units, sensors, and intelligent factories has become an essential part of manufacturing processes in Industry 4.0 [3]. However, the relevant data is often spread among different entities/locations within the same company or even controlled by different organizations. Furthermore, data security may prohibit sharing data. Thus, the processing and employment of such distributed data requires novel machine learning approaches. Federated learning (FL) [4] addresses these challenges by building models on decentralized data, thus keeping sensitive data on devices and avoiding central data storage.

FL is a distributed machine learning technique that allows multiple devices or systems to train a model together without sharing their raw data (Figure 1). Each device or system trains a local version of the model using its own data, and then the local models are aggregated to produce a global model. This can be particularly useful in the context of predictive maintenance or quality inspection, as it allows organizations to leverage data from multiple sources, such as sensor data from different machines or plants, while still maintaining the privacy and security of the individual data sets. Additionally, federated learning can enable organizations to improve the accuracy of their predictive models by training on a larger and more diverse set of data. Especially in an industrial context, data may not be shared between different entities for competitive reasons. In this case, federated learning allows participants to generate better models without disclosing their data to competitors. Overall, the potential for federated learning in industrial settings is to improve the accuracy of predictions while preserving the data privacy of each system.

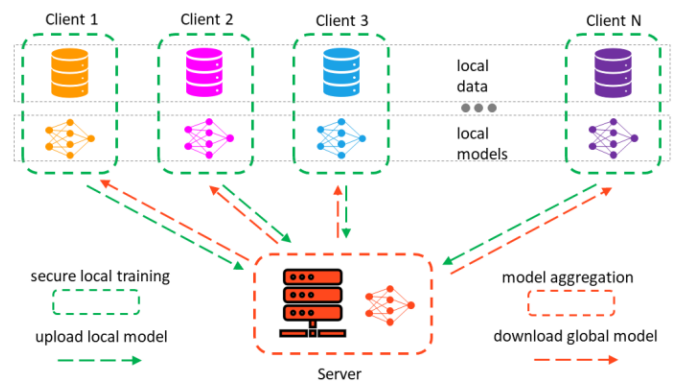

*Figure 1: Illustration of a federated learning system in which local models are trained by local data. The server centrally aggregates the local models to a global model.*

This research is supported by the Austrian Research Promotion Agency (FFG) under Contract No. 883855 INTERACTIVE.

XXX-X-XXXX-XXXX-X/XX/$XX.00 ©20XX IEEE

## A. Related Work

This section reviews relevant related work in FL, which addresses the application fields of predictive maintenance and quality control as its main focus.

Ge et al. [5] provide empirical research results for FL-based production line fault prediction. They constructed an FL support vector machine (SVM) and random forest (RF) approach for predicting the failures in a production line, compared them to centralized SVM and RF respectively, and achieved similar effectiveness between corresponding federated and central learning models.

Zhang et al. [6] used a combination of autoencoders and Siamese networks to detect faults in permanent magnet synchronous motors. They tested their proposed method with both centralized and federated learning, using sparse data. The results showed improved accuracy compared to other deep learning architectures. However, the study's generalizability is limited because their analysis is only based on a single, undisclosed dataset.

Zhang et al. [7] analyzed the effects of different types of data heterogeneity in machinery fault diagnostics. They introduced an FL averaging method combined with a dynamic validation scheme as well as self-supervision. They achieved promising performances on non-independent and identically distributed (i.i.d.) data. At the same time, they focused on modeling non-i.i.d.-class and non-i.i.d.-domain imbalance, e. g. by assigning different types of faulty states to different clients.

The work of Bemani and Björsell [8] deals with distributed ML for predictive maintenance applications. They proposed two federated algorithms: Federated support vector machine (FedSVM) for anomaly detection and federated long-short term memory (FedLSTM) for predicting engines' remaining useful life-time. Their results show that FedSVM and FedLSTM are comparable to centralized algorithms.

The work of Hiessl et al. [9] and Holly et al. [10] deals with data distribution shift and non i.i.d. real-world industrial data. While Hiessl focused on organizing clients into cohorts with similar data shifts, Holly focused on hyperparameter optimization in FL settings.

As far as we know, no studies have been conducted to evaluate multiple FL algorithms for imbalanced predictive maintenance or quality inspection data.

## B. Contributions

In this work, we address the question to what extent FL can replace centralized or local training without major losses in model accuracy. We explore four well-known aggregation strategies for FL – FedAvg [11], FedProx [12], qFedAvg [13], and FedYogi [14], because each aggregation scheme has its own trade-offs and limitations e. g. with respect to robustness and fairness.

The contributions of this article are summarized as follows:

- We evaluate four FL strategies (FedAvg, FedProx, qFedAvg, FedYogi) and determine the extent to which FL can replace centralized or local training without significant reductions in model accuracy.

- To enhance validity, the evaluation is carried out on three publicly available datasets as well as a new dataset dealing with quality inspection.

- We provide to the community a novel dataset (denoted as FLADI[1]) for federated learning in the context of quality inspection.

Note that we do not analyse the communication costs of the FL schemes in this work. The rest of this article is organized as follows: Section II describes the used methods and Section III the datasets. Section IV presents the experiments and discussion. Finally, Section V concludes this article.

## II. METHODS

Federated learning [15] is a decentralized machine learning approach where multiple clients train a model on their own data locally and periodically share the model updates with a central server. The number of clients can vary widely depending on the use case and available resources. The global update aggregates the local models to produce a global model. This allows for the training of models on large amounts of data that are distributed across multiple devices without the need to share the raw data with a central server, which can improve privacy and security. Additionally, FL has the potential to enable organizations to enhance the accuracy of their models by training on a more diverse range of data.

In practice, there are three main FL scenarios: horizontal federated learning (HFL), vertical federated learning (VFL), and federated transfer learning (FTL). HFL is applicable when clients share a similar feature space, but the objects are different. For example, there are some manufacturing units, that are spread geographically and produce similar products, thus measuring the same features during production. VFL is employed when two datasets share the same objects but differ in feature space. For instance, the same product is tracked over a sequence of stages of the production process and different metrics for this product are recorded at each stage. In FTL, the clients observe neither common features nor common objects. In this work, only HFL is considered. Data distribution among clients can have a significant impact on the performance of FL. For instance, if some clients have significantly different data distributions than others, the global model may not generalize well and be biased towards some clients. In FL, the training process is divided into two main components: local epochs and global rounds. Each client trains the model on its own local data for a certain number of local epochs. During these local epochs, the model parameters are updated based on the client's data. Afterwards, each client sends the updated model parameters to the central server, which determines a new global model. This process is repeated for a specified number of global rounds. The number of local epochs and global rounds can be adjusted to balance the trade-off between accuracy and communication overhead.

---

[1] https://www.stiwa.com/karriere/stiwa-educational-content

The central server orchestrates the training performed on the clients' side, collects the local updates centrally, and aggregates the updates into a global model. We explore four aggregation strategies - FedAvg, FedProx, qFedAvg, and FedYogi, as they put a different focus on optimization methods, robustness, and fairness:

- Federated averaging (FedAvg) [11] determines the global model by the weighted average of all model updates.

- FedProx [12] introduces a proximal term μ to restrict strong drifts of the global model towards any of the local ones by limiting the contribution size of the local updates to improve the stability of the training process.

- qFedAvg [13] is a fairness-driven federated optimization technique where clients with poor performance at a given time are dynamically assigned higher relative weights. This is accomplished through the fairness parameter q, which allows the user to control the degree of fairness by emphasizing clients with higher local losses.

- FedYogi [14] extends FedAvg with adaptivity options on both the client and server side. For optimization on the server, a global learning rate $\eta_g$ must be selected. Additionally, the parameter $\tau$ is used to manage the level of adaptivity.

We designed a multi-layer perceptron (MLP) model for binary classification tasks. The MLP has multiple hidden layers and two output neurons. Each hidden layer comprises several neurons (see Table 1) with ReLU activation and batch normalization techniques.

## III. DATASETS

Obtaining practical federated datasets can be difficult due to privacy and legal constraints [16]. To overcome this challenge, we have chosen three publicly accessible datasets that are suitable for a classification task with a high degree of class imbalance. Our selected datasets are single collections of data, and we simulate FL clients by dividing the data into partitions. Additionally, we performed an evaluation on a newly created real dataset named FLADI, which represents a realistic federated scenario, since the data can be assigned to individual clients.

### A. Public Datasets

The AI4I 2020 Predictive Maintenance Dataset (AI4I2020) [17][18] is a synthetic dataset, that aims to represent real-world industrial predictive maintenance data. The dataset comprises six features, consisting of two categorical variables for the product type and type of failure, as well as four numerical variables for process temperature, air temperature, rotational speed, and torque. We restrict our analysis to a binary task, focusing on the presence or absence of failure. The dataset consists of 10,000 data points, out of which 339 contain failures, resulting in a high-class imbalance with only 3.39% of samples having a positive (failure) class. We use 80% of the data for training, 10% for validation, and 10% for testing. The misclassification cost of a false negative is estimated to be 30 times the cost of a false positive.

Failures in Scania trucks dataset (Scania) [17][19] is a collection of various component failures in Scania trucks during normal operation. In this dataset, the positive class represents the failures that are related to the air pressure system, while the negative class covers the failures not associated with the air pressure system. The dataset contains 60,000 data points for training, and thereof only 1,000 records or 1.7% belong to the positive class. We use 20% of the training data for validation. For testing, an additional dataset is provided. It contains 16,000 records, of which 2.3% are positive. Each data point consists of 170 features, which represent operational measurements. The feature names are anonymized. The misclassification cost of a false negative is estimated to be 50 times of a false positive.

Backblaze [20] provides daily data and summary statistics about the performance of hard drives in their data centers (hereafter referred to as hard drive). The data includes several key pieces of information, such as the drive's serial number, model, capacity, 45 S.M.A.R.T. statistics in both raw and normalized forms, as well as a failure indicator. From these, we consider only a subset of features (S.M.A.R.T. metrics: 3, 5, 7, 187, 188, 190, 194, 197, 198, 199), which should be the most useful for predicting failures, as suggested by Amram et al. [21]. The dataset comprises various models of hard drives, with annualized failure rates that range from 0.5% to 2%. To have a binary classification problem, we consider only the short-term health of hard drives and aim to identify the faulty records at the same time as the measurements are taken. We use 381,000 records for training and 9,600 for validation as well as testing.

### B. Novel Dataset

The novel dataset denoted as **F**ederated **LeA**rning **D**ataset for Quality **I**nspection (FLADI) provided by the automotive industry manufacturer STIWA represents a real-world use case for the application of FL in the context of automation solutions. The dataset is derived from the production of gearbox components, in which prefabricated parts are first deformed and then joined using laser welding. During production, quality inspection engineers perform destructive testing on individual components. These detailed quality inspection results serve as the target variables. The dataset consists of 4,281 samples, grouped into four product variants (sample distribution: 807, 1198, 1166, 1110), with 138 features representing measurements such as position, force, and moment, and one quality metric. The feature names have been anonymized. Note, not all 138 features are assigned to each product variant, so the feature space per product variant differs slightly. These four product variants serve as individual clients in our FL scenario, simulating the production of similar products in different production facilities. The proportion of the data between acceptable and unacceptable product quality is approximately 80% to 20%. In this dataset, an excessive number of negative examples were included to provide the machine learning models with sufficient samples of failures. Business analysts estimated the maintenance costs to 5 units per false positive and 15 units per false negative detection. We provide further insights into the FLADI data set at the source referenced in Section 1.

## IV. EXPERIMENTS AND DISCUSSION

### A. Experiments

We defined a set of scenarios for our experiments. Unlike FLADI, the other datasets do not have the data assigned to the clients yet. We are conducting experiments with different numbers of clients as well as data distributions to investigate the impact on the FL performance. Table 1 displays the parameter values of the various experimental settings.

Regarding data distributions we consider three scenarios:
- i.i.d.: The data is randomly and evenly distributed among the clients.
- non-i.i.d. with feature distribution skew: To simulate clients with different feature distributions, we employ principal component analysis (PCA). For this, we calculate the principal components of our data and assign the observations to clients according to the value of their first principal component.
- non-i.i.d. with quantity skew: Clients are allocated with proportions of data according to the Dirichlet distribution as suggested by Yurochkin et al. [22].

Classifying unbalanced data generally leads to poor prediction accuracy and low generalization ability of the model [23]. We address this issue twofold – by adjusting the weights of classes in the cost function and by data resampling. For the latter, we apply the Synthetic Minority Oversampling Technique (SMOTE) [24]. Instead of simply duplicating the existing minority samples, SMOTE is used to generate new synthetic samples from the minority distribution. For each minority instance, a neighborhood of the k closest minority instances is considered, and new instances are created by moving in one or several directions towards the nearest neighbors. To tackle the data imbalance issue of the AI4I2020 and Scania datasets, we employ SMOTE oversampling technique. We generate minority samples until the total number of failures is 20% of the majority class size.

*Table 1: Model and experimental settings.*

| Parameter | AI4I2020 | Scania | Hard Drive | FLADI |
|---|---|---|---|---|
| MLP (number of neurons) | 50,20,10 | 200,100, 50 | 400, 200, 100, 50 | 35,15, 15 |
| Learning rate | 1e-05 | 1e-04 | 1e-05 | 1e-04 |
| Batch size | 128 | 128 | 1024 | 75 |
| # clients | 5,10,15 | 10,20,30 | 3,10,20 | 4 |
| # local epochs | 10 | 10 | 5 | 20 |
| # global rounds | 100 | 100 | 50 | 500 |
| Proximal term $\mu$ (FedProx) | 0.1 | 0.001 | 0.1 | 0.05 |
| Fairness parameter q (qFedAvg) | 5e-12 | 5e-05 | 1 | 5e-06 |
| Global learning rate $\eta_g$ (FedYogi) | 0.1 | 0.01 | 0.001 | 0.1 |
| Degree of adaptivity $\tau$ (FedYogi) | 1e-05 | 1e-05 | 0.001 | 1e-05 |
| Beta score (F-beta) | 30 | 50 | 1 | 3 |

We selected a model architecture for all our experiments to make results comparable (Tab. 1). We compute the loss based on the cross entropy weighted by classes, apply dropout, and optimize our networks using the ADAM algorithm. The definitive network architecture and training parameters, such as learning rate, were optimized using the centralized learning scenario. For each data set, we conducted all our experiments with the same randomly initialized model and parameter values as specified in Table 1.

We evaluated the models by calculating the F-beta scores [25], in which beta is a weighted harmonic mean of precision and recall. Beta values of less than one assign more importance to precision, while beta values greater than one assign more importance to recall. By setting the parameter beta, we can control the relative importance (costs) of the different types of prediction mistakes. We set beta equal to the ratio of the costs of false positives versus false negatives (see Table 1). Further, we also determined the fairness of the FL schemes by calculating the entropy of clients' individual F-beta scores [26].

### B. Discussion

All results can be found in Table 2. As expected, in most cases, central learning leads to the best results and an increased number of clients in the local learning scenario leads to a performance drop. A noteworthy exception is the Scania non-i.i.d. scenario with feature distribution skew, where local learning as well as all four FL strategies with several clients outperformed central learning. We suspect that this unusual behavior is triggered by the PCA preprocessing of the data.

Our initial hypothesis that FL will always outperform local learning, however, could not be confirmed for all data sets. While in the Hard Drive dataset an overall improvement compared to local learning was observed through all data distributions, no improvement could be seen in the AI4I2020 or FLADI datasets. Interestingly, in both datasets, central learning clearly outperformed local learning. In the FLADI data set, each client has a slightly different feature space due to different product variants, which may have caused the model to not converge well during the federation.

In general, the performance of models trained using FL degrades as the number of clients increases. For instance, the F-beta score of FedAvg calculated on AI4I2020 decreases from 0.93 (5 clients) to 0.75 (15 clients). The situation is similar for the distribution of the data. As a rule, the results of the FL models are worse for the non-i.i.d. data distributions. The comparison of four FL strategies to update the global model shows that FedAvg and FedProx outperform the other two fairness driven algorithms.

In our experiments, both qFedAvg and FedYogi either deliver poor performance (e. g. for AI4I2020 or hard drive) or the improvement of the fairness is neglectable. For instance, the fairness values of the models trained on FLADI are: 1.9819 (FedAvg), 1.9857 (FedProx), 1.9901 (qFedAvg) and 1.9838 (FedYogi) and the differences of the entropy are very small. For further theoretical analysis of our results, we refer to the master's thesis by Viktorija Pruckovskaja [27]. Our experiments show an influence of the data distribution on the performance, and thus the generalizability is limited to use cases with similar distributions.

Our paper presents a novel contribution to the field of federated learning by providing FLADI, a real-world dataset that represents the complexities and diversity of the industry.

Unlike typical FL scenarios, this dataset showcases the reality of having varying features among different product variants, which is a significant challenge in the industry. This dataset will play a crucial role in advancing the field by providing a more realistic and applicable scenario for FL in the industry.

## V. CONCLUSIONS

Our key objective was to evaluate whether FL might be a good alternative to traditional central or local training schemes in industrial settings. Therefore, we implemented four well-known FL aggregation schemes and evaluated the different training strategies on four different datasets, including one novel real-world FL dataset which we provide to the research community to foster future developments in FL.

Our analysis showed that in respective industrial datasets, the benefit of current FL approaches may depend on the data itself and that federated learning may not be suitable for all use cases. One possible cause for this may be the data distribution shift between clients. Another common challenge in industrial settings are heterogeneous feature sets over FL clients, as demonstrated by the FLADI data set.

## ACKNOWLEDGMENT

We especially want to thank Stefan Stricker from STIWA Automation for fruitful discussions and his contributions to the FLADI data set.

*Table 2: Calculated F-beta scores of the four different datasets with various data distributions, number of clients and training methods.*

| Dataset | Data Distribution | # clients | Training method | | | | | |
|---|---|---|---|---|---|---|---|---|
| | | | **Central** | **Local** | **FedAvg** | **FedProx** | **qFedAvg** | **FedYogi** |
| AI4I2020 | i.i.d | 5 | 0,95 | 0,89 | 0,93 | 0,93 | 0,48 | 0,67 |
| | | 10 | 0,95 | 0,80 | 0,77 | 0,78 | 0,39 | 0,39 |
| | | 15 | 0,95 | 0,78 | 0,75 | 0,75 | 0,42 | 0,42 |
| | Non-i.i.d. with feature distribution skew | 5 | 0,95 | 0,92 | 0,74 | 0,74 | 0,49 | 0,50 |
| | | 10 | 0,95 | 0,88 | 0,66 | 0,66 | 0,48 | 0,50 |
| | | 15 | 0,95 | 0,89 | 0,66 | 0,66 | 0,53 | 0,50 |
| | Non-i.i.d. with quantity skew | 5 | 0,95 | 0,79 | 0,80 | 0,80 | 0,39 | 0,53 |
| | | 10 | 0,95 | 0,82 | 0,80 | 0,80 | 0,43 | 0,43 |
| | | 15 | 0,95 | 0,73 | 0,70 | 0,70 | 0,39 | 0,61 |
| Scania | i.i.d | 10 | 0,95 | 0,95 | 0,96 | 0,96 | 0,94 | 0,96 |
| | | 20 | 0,95 | 0,93 | 0,95 | 0,95 | 0,94 | 0,96 |
| | | 30 | 0,95 | 0,92 | 0,93 | 0,94 | 0,94 | 0,96 |
| | Non-i.i.d. with feature distribution skew | 10 | 0,95 | 0,99 | 0,99 | 0,99 | 0,99 | 0,99 |
| | | 20 | 0,95 | 0,99 | 0,99 | 0,99 | 0,99 | 0,99 |
| | | 30 | 0,95 | 0,99 | 0,99 | 0,99 | 0,99 | 0,99 |
| | Non-i.i.d. with quantity skew | 10 | 0,95 | 0,92 | 0,95 | 0,95 | 0,94 | 0,96 |
| | | 20 | 0,95 | 0,92 | 0,94 | 0,93 | 0,93 | 0,96 |
| | | 30 | 0,95 | 0,87 | 0,91 | 0,91 | 0,92 | 0,94 |
| Hard drive | i.i.d | 3 | 0,73 | 0,73 | 0,74 | 0,73 | 0,55 | 0,50 |
| | | 10 | 0,73 | 0,69 | 0,74 | 0,73 | 0,55 | 0,50 |
| | | 20 | 0,73 | 0,65 | 0,73 | 0,71 | 0,54 | 0,50 |
| | Non-i.i.d. with feature distribution skew | 3 | 0,73 | 0,70 | 0,69 | 0,68 | 0,54 | 0,50 |
| | | 10 | 0,73 | 0,69 | 0,74 | 0,73 | 0,55 | 0,50 |
| | | 20 | 0,73 | 0,65 | 0,73 | 0,72 | 0,54 | 0,50 |
| | Non-i.i.d. with quantity skew | 3 | 0,73 | 0,74 | 0,73 | 0,74 | 0,56 | 0,50 |
| | | 10 | 0,73 | 0,70 | 0,76 | 0,74 | 0,55 | 0,49 |
| | | 20 | 0,73 | 0,66 | 0,73 | 0,70 | 0,56 | 0,50 |
| FLADI | Non-i.i.d. with feature distribution and quantity skew | 4 | 0,70 | 0,63 | 0,63 | 0,61 | 0,64 | 0,60 |